\documentclass{article}

\PassOptionsToPackage{numbers, compress}{natbib}

\usepackage[final]{neurips_2025}

\usepackage[utf8]{inputenc} 
\usepackage[T1]{fontenc}    
\usepackage{hyperref}       
\usepackage{url}            
\usepackage{booktabs}       
\usepackage{amsfonts}       
\usepackage{nicefrac}       
\usepackage{microtype}      
\usepackage[dvipsnames]{xcolor}         

\usepackage{wrapfig}
\usepackage{float}

\usepackage{etoc}

\usepackage{amsmath}
\usepackage{amsthm}
\usepackage{amsfonts}
\usepackage{amssymb}
\usepackage{mathtools}
\usepackage{tcolorbox}

\newcommand{\matr}[1]{\mathbf{#1}}

\DeclareMathOperator{\LN}{LN}
\DeclareMathOperator{\DA}{DA}
\DeclareMathOperator{\MSD}{MSD}

\usepackage[capitalize,noabbrev]{cleveref}

\hypersetup{
  colorlinks,
  citecolor=orange,
  linkcolor=blue,
  urlcolor=blue
}

\usepackage[textsize=tiny]{todonotes}

\usepackage{tikz}
\usetikzlibrary{calc}
\tikzstyle{mynode}=[thick,draw=black,circle,minimum size=15]

\theoremstyle{plain}
\newtheorem{theorem}{Theorem}

\theoremstyle{definition}

\theoremstyle{remark}

\theoremstyle{definition}

\newenvironment{mythm}[1]{%
  \renewcommand\thetheorem{#1}
  \theorem
}{\endtheorem}

\title{A Simple Generalisation of the Implicit Dynamics of In-Context Learning}

\author{%
  Francesco Innocenti\normalfont\textsuperscript{$1, 2, 3$} \\
  \And
  El Mehdi Achour\normalfont\textsuperscript{$4$} \\
  \AND
  {\normalfont\textsuperscript{$1$}MRC Brain Network Dynamics Unit, University of Oxford, UK}\\
  {\normalfont\textsuperscript{$2$}MRC CoRE in Restorative Neural Dynamics, UK}\\
  {\normalfont\textsuperscript{$3$}University of Sussex, Brighton, UK}\\
  {\normalfont\textsuperscript{$4$}University Mohammed VI Polytechnic, College of Computing, Rabat, Morocco}\\
  Correspondence to: \texttt{francesco.innocenti@ndcn.ox.ac.uk}
}

\begin{document}

\maketitle

\begin{abstract}
  In-context learning (ICL) refers to the ability of a model to learn new tasks from examples in its input without any parameter updates. In contrast to previous theories of ICL relying on toy models and data settings, recently it has been shown that an abstraction of a transformer block can be seen as implicitly updating the weights of its feedforward network according to the context \cite{dherin2025learning}. Here, we provide a simple generalisation of this result for (i) all sequence positions beyond the last, (ii) any transformer block beyond the first, and (iii) more realistic residual blocks including layer normalisation. We empirically verify our theory on simple in-context linear regression tasks and investigate the relationship between the implicit updates related to different tokens within and between blocks. These results help to bring the theory of \cite{dherin2025learning} even closer to practice, with potential for validation on large-scale models.
\end{abstract}

\section{Motivation and main result}
\label{main-result}

Large-scale pretrained models show a remarkable emergent ability to learn new tasks from examples in their input without any fine-tuning or parameter updates. This ``in-context learning'' (ICL) capability was first noted for GPT-3 \cite{brown2020language} and has more recently also been shown by large vision models \cite{bar2022visual}. For this reason, there has been increasing interest in understanding the mechanisms behind ICL \cite{dong2022survey, zhou2023mystery}, using both empirical and theoretical approaches. While previous theoretical analyses of ICL have relied on simplified models and data settings, recently \cite{dherin2025learning} showed that an abstraction of a transformer block—consisting of a ``context-aware'' layer such as self-attention \cite{vaswani2017attention} and a multi-layer perceptron (MLP)—has the implicit effect of modifying the MLP weights according to the context.

However, among other limitations, the analysis of \cite{dherin2025learning} applies only to the last token and the first transformer block, and their extension to blocks with skip connections does not exactly correspond to the standard Pre-LayerNorm (LN) transformer architecture used in practice \cite{wang2019learning, xiong2020layer}. Our main contribution is to generalise the main result of \cite{dherin2025learning} in all these respects, namely for any token, block and more accurate residual blocks including layer normalisation.

Following their setup, we define a \textit{contextual layer} $\matr{A}: \mathbb{R}^{d \times N} \rightarrow \mathbb{R}^{d \times N}$ as any layer such as self-attention that can process a $d$-dimensional input sequence of any length $N$. We ignore multiple sequence batches for simplicity. The contextual layer can take as input either a single query vector $\matr{A}(\mathbf{x})$ with $\mathbf{x} \in \mathbb{R}^{d}$, or also a context sequence in addition to the query $\matr{A}(\matr{C}, \mathbf{x})$, with $\matr{C} \in \mathbb{R}^{d \times (N-1)}$. A \textit{contextual block} is then defined as the stacking of a contextual layer with an MLP, $\matr{T}_{\matr{W}}(\cdot) = \matr{W}' \big( \sigma(\matr{W} \matr{A}(\cdot) + \mathbf{b}) \big) + \mathbf{b}' \in \mathbb{R}^{d \times N}$ with weights $(\matr{W}, \matr{W}')$, biases $(\mathbf{b}, \mathbf{b}')$ and activation function $\sigma$. 

Based on these definitions, a simplified version of the main result of \cite{dherin2025learning} can be stated as follows:\footnote{The more general version of the theorem considers any subset of the context $\matr{Y} \subset \matr{C}$ that may modify $\Delta \matr{W}_N(\matr{Y})$, which we will ignore for simplicity.}
\begin{equation}
    \matr{T}_{\matr{W}}(\matr{C}, \mathbf{x})_{(N)} = \matr{T}_{\matr{W} + \Delta \matr{W}_N(\matr{C})}(\mathbf{x}), \quad \Delta \matr{W}_N(\matr{C}) = \frac{(\matr{W} \Delta \matr{A}_{(N)}) \matr{A}(\mathbf{x})^T}{||\matr{A}(\mathbf{x})||^2}
    \label{eq:dherin-result}
\end{equation}
where $\Delta \matr{A}_{(N)} = \matr{A}(\matr{C}, \mathbf{x})_{(N)} - \matr{A}(\mathbf{x})$ is the difference in the contextual layer's prediction of the last token with and without context, and $N$ indexes the last sequence element, which is left implicit in \cite{dherin2025learning}. Eq. \ref{eq:dherin-result} shows that the last-token prediction of a contextual (e.g. transformer) block taking some context and query as input (LHS) is equivalent to that of the same block with only the query as input and the first weight matrix of the MLP updated by the context (RHS). Notably, the implicit weight update $\Delta \matr{W}_N(\matr{C})$ is of rank one, as the outer product of a column vector and a row vector.
\begin{figure}[t]
    \begin{center}
        \centerline{\includegraphics[width=0.9\textwidth]{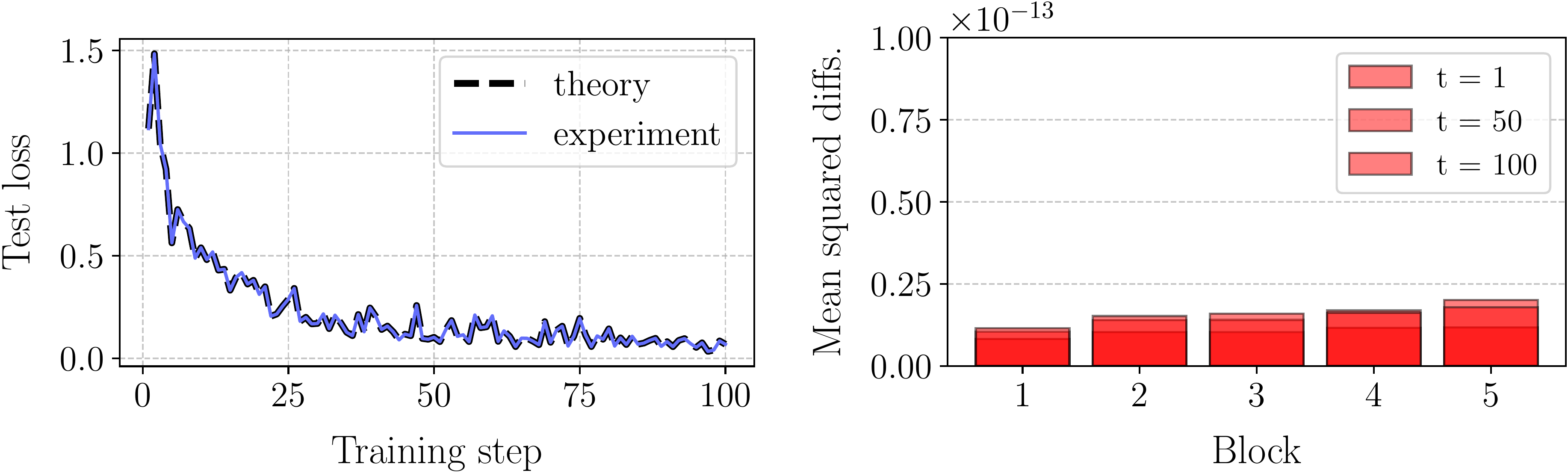}}
        \caption{\textbf{Empirical verification of Theorem \ref{thm1} for in-context linear regression.} (\textit{Left}) Test losses of a 5-layer transformer trained to solve linear regression tasks in context (see \S \ref{exp-details} for details). The empirical and theoretical losses were computed using the left- and right-hand side of Eq. \ref{eq:main}, respectively, for the last block $\ell = 5$ and token $i=N$. (\textit{Right}) Mean squared differences between the theoretical and empirical predictions (see Eq. \ref{eq:mse-metric}) of every block at different training steps $t$. Results were consistent across different random seeds.}
        \label{fig:theory-verification}
    \end{center}
    \vskip -0.25in
\end{figure}

Our generalisation of Eq. \ref{eq:dherin-result}, given in the following theorem, shows that the prediction of \textit{any} token $i$ by \textit{any} contextual block $\ell$ with more realistic skip connections including Pre-LN (see \S \ref{thm1-proof} for details) is equivalent to that of the same block with only the previous query as input and specific MLP parameters updated by the context. For any block other than the first, we can think of the inputs $(\matr{C}_\ell, \mathbf{x}_\ell)$ as ``refined'' versions of the original context and query. 
\begin{tcolorbox}[width=\linewidth, sharp corners=all, colback=white!95!black, colframe=white!95!black]
    \begin{mythm}{1}[]\label{thm1}
        Consider a contextual block $\matr{T}_{\matr{W}, \mathbf{b}'}^\ell$ with skip connections, Pre-LN (as in Eq. \ref{eq:pre-ln-block}) and input $(\matr{C}_\ell, \mathbf{x}_\ell)$. Then, the following equality holds (see \S \ref{thm1-proof} for proof):
        \begin{equation}
            \matr{T}^\ell_{\matr{W}, \mathbf{b}'}(\matr{C}_\ell, \mathbf{x}_\ell)_{(i)} = \matr{T}^\ell_{\matr{W} + \Delta \matr{W}_i(\matr{C}), \mathbf{b}' + \Delta \mathbf{b}_i'(\matr{C})}(\mathbf{x}_\ell),
        \label{eq:main}
        \end{equation}
        where the MLP updates of the first weight matrix and the last layer's biases are given in Eqs. \ref{eq:weight-update-ln} and \ref{eq:bias-update-ln}, respectively. The weight update (Eq. \ref{eq:weight-update-ln}) is of rank one as in \cite{dherin2025learning}.
    \end{mythm}
\end{tcolorbox}
Following \cite{dherin2025learning} and other previous works \cite{garg2022can, zhang2024trained}, we verified our theory on the well-defined problem of in-context linear regression by testing multi-layer transformers to predict sequences of linear functions that were not previously seen during training (see \S \ref{exp-details} for details). Figure \ref{fig:theory-verification} shows an excellent match between the theory and experiment. Additional analyses of the implicit weight updates related to different token positions within and between blocks are included in Appendix \ref{appendix}.

To conclude, our results help to bring the theory of \cite{dherin2025learning} even closer to practice, potentially allowing for validation on large-scale models. In particular, it would be interesting to analyse the implicit weight updates of models trained on language, which our generalisation enables. Our work is still limited by considering one step of token generation, and it could be important to study ICL settings where the answer is itself a sequence of tokens.



\medskip

\bibliography{references}

\begin{thebibliography}{10}

\bibitem{ba2016layer}
J.~L. Ba, J.~R. Kiros, and G.~E. Hinton.
\newblock Layer normalization.
\newblock {\em arXiv preprint arXiv:1607.06450}, 2016.

\bibitem{bar2022visual}
A.~Bar, Y.~Gandelsman, T.~Darrell, A.~Globerson, and A.~Efros.
\newblock Visual prompting via image inpainting.
\newblock {\em Advances in Neural Information Processing Systems}, 35:25005--25017, 2022.

\bibitem{brown2020language}
T.~Brown, B.~Mann, N.~Ryder, M.~Subbiah, J.~D. Kaplan, P.~Dhariwal, A.~Neelakantan, P.~Shyam, G.~Sastry, A.~Askell, et~al.
\newblock Language models are few-shot learners.
\newblock {\em Advances in neural information processing systems}, 33:1877--1901, 2020.

\bibitem{dherin2025learning}
B.~Dherin, M.~Munn, H.~Mazzawi, M.~Wunder, and J.~Gonzalvo.
\newblock Learning without training: The implicit dynamics of in-context learning.
\newblock {\em arXiv preprint arXiv:2507.16003}, 2025.

\bibitem{dong2022survey}
Q.~Dong, L.~Li, D.~Dai, C.~Zheng, J.~Ma, R.~Li, H.~Xia, J.~Xu, Z.~Wu, T.~Liu, et~al.
\newblock A survey on in-context learning.
\newblock {\em arXiv preprint arXiv:2301.00234}, 2022.

\bibitem{garg2022can}
S.~Garg, D.~Tsipras, P.~S. Liang, and G.~Valiant.
\newblock What can transformers learn in-context? a case study of simple function classes.
\newblock {\em Advances in neural information processing systems}, 35:30583--30598, 2022.

\bibitem{kingma2014adam}
D.~P. Kingma and J.~Ba.
\newblock Adam: A method for stochastic optimization.
\newblock {\em arXiv preprint arXiv:1412.6980}, 2014.

\bibitem{vaswani2017attention}
A.~Vaswani, N.~Shazeer, N.~Parmar, J.~Uszkoreit, L.~Jones, A.~N. Gomez, {\L}.~Kaiser, and I.~Polosukhin.
\newblock Attention is all you need.
\newblock {\em Advances in neural information processing systems}, 30, 2017.

\bibitem{wang2019learning}
Q.~Wang, B.~Li, T.~Xiao, J.~Zhu, C.~Li, D.~F. Wong, and L.~S. Chao.
\newblock Learning deep transformer models for machine translation.
\newblock {\em arXiv preprint arXiv:1906.01787}, 2019.

\bibitem{xiong2020layer}
R.~Xiong, Y.~Yang, D.~He, K.~Zheng, S.~Zheng, C.~Xing, H.~Zhang, Y.~Lan, L.~Wang, and T.~Liu.
\newblock On layer normalization in the transformer architecture.
\newblock In {\em International conference on machine learning}, pages 10524--10533. PMLR, 2020.

\bibitem{zhang2024trained}
R.~Zhang, S.~Frei, and P.~L. Bartlett.
\newblock Trained transformers learn linear models in-context.
\newblock {\em Journal of Machine Learning Research}, 25(49):1--55, 2024.

\bibitem{zhou2023mystery}
Y.~Zhou, J.~Li, Y.~Xiang, H.~Yan, L.~Gui, and Y.~He.
\newblock The mystery of in-context learning: A comprehensive survey on interpretation and analysis.
\newblock {\em arXiv preprint arXiv:2311.00237}, 2023.

\end{thebibliography}


\newpage
\appendix

\section{Appendix} 
\label{appendix}

\localtableofcontents
\renewcommand{\thefigure}{A.\arabic{figure}}
\setcounter{figure}{0}


\subsection{Proofs and derivations}
\label{proofs}

\subsubsection{Extension to all sequence positions}
\label{any-seq-pos-result}

Generalising the main result of \cite{dherin2025learning} (Theorem 2.2) to all output sequence positions (including the last) is simply a matter of indexing. As made explicit by our indexing in Eq. \ref{eq:dherin-result}, \cite{dherin2025learning} focus only on the last-token prediction of the contextual block $\matr{T}_{\matr{W}}(\matr{C}, \mathbf{x})_{(N)}$. We can therefore relax the result by simply considering any token position $i$
\begin{equation}
    \matr{T}_{\matr{W}}(\matr{C}, \mathbf{x})_{(i)} = \matr{T}_{\matr{W} + \Delta \matr{W}_i(\matr{C})}(\mathbf{x}), \quad \Delta \matr{W}_i(\matr{C}) = \frac{(\matr{W} \Delta \matr{A}_{(i)}) \matr{A}(\mathbf{x})^T}{||\matr{A}(\mathbf{x})||^2}.
    \label{eq:one-block-any-pos}
\end{equation}
where one only needs to index the output of the contextual layer $\Delta \matr{A}_{(i)} = \matr{A}(\matr{C}, \mathbf{x})_{(i)} - \matr{A}(\mathbf{x})$. Note that different sequence positions will be associated with different weight updates $\Delta \matr{W}_i(\matr{C})$ and that each update retains rank 1. This result can also be rewritten in matrix form by stacking all the weight updates related to different positions into a single matrix $\matr{B}$ 
\begin{equation}
    \matr{T}_{\matr{W}}(\matr{C}, \mathbf{x}) = \matr{T}_{\matr{B} + \Delta \matr{B}(\matr{C})}(\mathbf{x})
    \label{eq:one-block-any-pos-matrix-form}
\end{equation}
where the new matrix and its update are
\begin{align}
    \matr{B}
    = \left(\begin{array}{c}
         \matr{W} \\
         \matr{W} \\
         \vdots \\
         \matr{W}
    \end{array}\right) \in \mathbb{R}^{(hN) \times d} \quad \text{and} \quad \Delta \matr{B}(\matr{C}) = 
    \left(\begin{array}{c}
         \Delta \matr{W}_1(\matr{C}) \\
         \Delta \matr{W}_2(\matr{C}) \\
         \vdots \\
         \Delta \matr{W}_N(\matr{C})
    \end{array}\right) \in \mathbb{R}^{(hN) \times d}
    \label{eq:stacked-weight-matrices}
\end{align}
with $\matr{W} \in \mathbb{R}^{h \times d}$. It is straightforward to show that the rank of this update matrix $\Delta \matr{B}(\matr{C})$ is also 1. In particular, $\operatorname{rank}\left(\begin{array}{c}
     \Delta \matr{W}_1(\matr{C}) \\
     \Delta \matr{W}_2(\matr{C}) \\
     \vdots \\
     \Delta \matr{W}_N(\matr{C})
\end{array}\right) \leq N$. However, all the $\Delta \matr{W}_i(\matr{C})$ can be written by definition as the outer product of a column vector and a row vector $\mathbf{u}_i\mathbf{v}^T$, where $\mathbf{u}_i = \matr{W} \Delta \matr{A}_{(i)} \in \mathbb{R}^{h}$ and the same $\mathbf{v} = \matr{A}(\mathbf{x}) \in \mathbb{R}^{d}$ for all $i$. Hence 
\begin{align*}
\Delta \matr{B}(\matr{C}) = \left(\begin{array}{c}
     \Delta \matr{W}_1(\matr{C}) \\
     \Delta \matr{W}_2(\matr{C}) \\
     \vdots \\
     \Delta \matr{W}_N(\matr{C})
\end{array}\right) = \left(\begin{array}{c}
     \mathbf{u}_1 \mathbf{v}^T \\
     \mathbf{u}_2 \mathbf{v}^T \\
     \vdots \\
     \mathbf{u}_N \mathbf{v}^T
\end{array}\right) = \left(\begin{array}{c}
     \mathbf{u}_1 \\
     \mathbf{u}_2 \\
     \vdots \\
     \mathbf{u}_N
\end{array}\right)\mathbf{v}^T,
\end{align*}
which shows that $\operatorname{rank}(\Delta \matr{B}(\matr{C})) = 1 $.

\subsubsection{Extension to any contextual block}
\label{any-block-result}

Similar to the previous extension to all sequence positions (\S \ref{any-seq-pos-result}), the main result of \cite{dherin2025learning} can be generalised to any contextual block beyond the first one by simple iterative application. For any block $\ell$ other than the first, we can do this by thinking of their inputs $(\matr{C}_\ell, \mathbf{x}_\ell)$ as refined versions of the original (unprocessed) input context and query $(\matr{C}_1, \mathbf{x}_1)$. Hence
\begin{equation}
    \matr{T}^\ell_{\matr{W}}(\matr{C}_\ell, \mathbf{x}_\ell)_{(i)} = \matr{T}^\ell_{\matr{W} + \Delta \matr{W}_i(\matr{C}_\ell)}(\mathbf{x}_\ell), \quad \Delta \matr{W}_i(\matr{C}_\ell) = \frac{(\matr{W} \Delta \matr{A}^\ell_{(i)}) \matr{A}^\ell(\mathbf{x}_\ell)^T}{||\matr{A}^\ell(\mathbf{x}_\ell)||^2}.
    \label{eq:any-block-any-pos}
\end{equation}
where $\matr{T}_{\matr{W}}^\ell$ and $\matr{A}^\ell$ indicate the $\ell$th contextual block and layer, respectively. Note that, as expected, Eq. \ref{eq:any-block-any-pos} simplifies to Eq. \ref{eq:dherin-result} for $i = N$ and $\ell = 1$.

\subsubsection{Extension to any block with more accurate skip connections}
\label{more-accurate-skips-results}

Motivated by the Pre-LN architecture \cite{wang2019learning, xiong2020layer}, \cite{dherin2025learning} consider blocks with the following skip connections:
\begin{equation}
    \matr{T}(\matr{C}, \mathbf{x})_{(N)} = \mathbf{x} + \matr{A}(\matr{C}, \mathbf{x})_{(N)} + \matr{W}'\sigma \big( \matr{W} \matr{A}(\matr{C}, \mathbf{x})_{(N)} + \mathbf{b} \big) + \mathbf{b}',
    \label{eq:dherin-skips}
\end{equation}
However, this fails to input the full contextual layer's output into the MLP, specifically the input skip $\mathbf{x}$. The more exact block structure would be
\begin{equation}
    \matr{T}(\matr{C}, \mathbf{x})_{(N)} = \mathbf{x} + \matr{A}(\matr{C}, \mathbf{x})_{(N)} + \matr{W}'\sigma \Big( \matr{W} \big( \matr{A}(\matr{C}, \mathbf{x})_{(N)} + \textcolor{Green}{\mathbf{x}} \big) + \mathbf{b} \Big) + \mathbf{b}'
    \label{eq:correct-skips-last-token}
\end{equation}
where now the MLP is also fed the input skip $\textcolor{Green}{\mathbf{x}}$. To lay the groundwork for the proof of Theorem \ref{thm1}, we proceed in 3 main steps. First, using the same logic as in \S \ref{any-seq-pos-result}, we extend Eq. \ref{eq:correct-skips-last-token} to any token position $i$
\begin{equation}
    \matr{T}(\matr{C}, \mathbf{x})_{(i)} = (\matr{C}, \mathbf{x})_{(i)} + \matr{A}(\matr{C}, \mathbf{x})_{(i)} + \matr{W}'\sigma \Big( \matr{W} \big( \matr{A}(\matr{C}, \mathbf{x})_{(i)} + (\matr{C}, \mathbf{x})_{(i)} \big) + \mathbf{b} \Big) + \mathbf{b}'
    \label{eq:correct-skips-any-pos}
\end{equation}
where note that the input skip is equal to the query vector $(\matr{C}, \mathbf{x})_{(i)} = \mathbf{x}$ for the last position $i = N$. Second, we prove Theorem \ref{thm1} without layer normalisation for the first block, which can be stated as follows:
\begin{equation}
    \matr{T}_{\matr{W}, \mathbf{b}'}(\matr{C}, \mathbf{x})_{(i)} = \matr{T}_{\matr{W}_i(\matr{C}), \mathbf{b}_i'(\matr{C})}(\mathbf{x})
    \label{eq:main-first-block-only}
\end{equation}
where the updates of the first MLP weight matrix and the biases of the last layer are given by 
\begin{align}
    \Delta \matr{W}_i(\matr{C}) &= \frac{\big( \matr{W} (\Delta \matr{A}_{(i)} + \Delta \mathbf{z}_{(i)}) \big) \big( \matr{A}(\mathbf{x}) + \mathbf{x} \big)^T}{||\matr{A}(\mathbf{x}) + \mathbf{x}||^2} \label{eq:skips-weight-update-first-block} \quad \text{and} \\
    \Delta \matr{b}'_i(\matr{C}) &= \Delta \matr{A}_{(i)} + \Delta \mathbf{z}_{(i)}, \label{eq:skips-bias-update-first-block}
\end{align}
with $\Delta \mathbf{z}_{(i)} = (\matr{C}, \mathbf{x})_{(i)} - \mathbf{x}$ as the difference between any input element and the query. The result now follows by direct computation as in \cite{dherin2025learning}. Let $\matr{W}_i(\matr{C}) = \matr{W} + \Delta \matr{W}_i(\matr{C})$ and $\mathbf{b}_i'(\matr{C}) = \mathbf{b}' + \Delta \mathbf{b}_i'(\matr{C})$. Then by definition, the right-hand side of Eq. \ref{eq:main-first-block-only} is
\begin{align}
    \matr{T}_{\matr{W}_i(\matr{C}), \mathbf{b}_i'(\matr{C})}(\mathbf{x}) 
    &= \mathbf{x} + \matr{A}(\mathbf{x}) + \matr{W}'\sigma \Big( \big( \matr{W} + \Delta \matr{W}_i(\matr{C}) \big) \big( \matr{A}(\mathbf{x}) + \mathbf{x}) \big) + \mathbf{b} \Big) + \mathbf{b}' + \Delta \mathbf{b}_i'(\matr{C}) \\
    &= \mathbf{x} + \matr{A}(\mathbf{x}) + \Delta \mathbf{b}_i'(\matr{C}) + \matr{W}'\sigma \Big( \big( \matr{W} + \Delta \matr{W}_i(\matr{C}) \big) \big( \matr{A}(\mathbf{x}) + \mathbf{x}) \big) + \mathbf{b} \Big) + \mathbf{b}',
\end{align}
where the second line simply moves the update of the last layer's biases for later convenience. Substituting $\Delta \matr{W}_i(\matr{C})$ (Eq. \ref{eq:skips-weight-update-first-block}) and using the fact that $\frac{\mathbf{x}^T}{||\mathbf{x}||^2}\mathbf{x} = 1$, we obtain
\begin{align}
    \Delta \matr{W}_i(\matr{C}) \big( \matr{A}(\mathbf{x}) + \mathbf{x}) \big) &= \frac{\big( \matr{W} (\Delta \matr{A}_{(i)} + \Delta \mathbf{z}_{(i)}) \big) \big( \matr{A}(\mathbf{x}) + \mathbf{x} \big)^T}{||\big( \matr{A}(\mathbf{x}) + \mathbf{x} \big)||^2} \big( \matr{A}(\mathbf{x}) + \mathbf{x} \big) \\
    &= \matr{W} (\Delta \matr{A}_{(i)} + \Delta \mathbf{z}_{(i)}),
\end{align}
which gives
\begin{align}
    \matr{T}_{\matr{W}_i(\matr{C}), \mathbf{b}_i'(\matr{C})}(\mathbf{x}) 
    = \mathbf{x}
    + \matr{A}(\mathbf{x}) + \Delta \mathbf{b}_i'(\matr{C})
    + \matr{W}'\sigma \Big( \matr{W} \big( \matr{A}(\mathbf{x}) + \mathbf{x} + \Delta \matr{A}_{(i)} + \Delta \mathbf{z}_{(i)} \big) + \mathbf{b} \Big) + \mathbf{b}'.
\end{align}
By the above definitions, we have that $\matr{A}(\mathbf{x}) + \Delta \matr{A}_{(i)} = \matr{A}(\matr{C}, \mathbf{x})_{(i)}$ and $\mathbf{x} + \Delta \mathbf{z}_{(i)} = (\matr{C}, \mathbf{x})_{(i)}$. Hence,
\begin{align}
    \matr{T}_{\matr{W}_i(\matr{C}), \mathbf{b}_i'(\matr{C})}(\mathbf{x}) 
    = \mathbf{x}
    + \matr{A}(\mathbf{x}) + \Delta \mathbf{b}_i'(\matr{C})
    + \matr{W}'\sigma \Big( \matr{W} \big( \matr{A}(\matr{C}, \mathbf{x})_{(i)} + (\matr{C}, \mathbf{x})_{(i)} \big) + \mathbf{b} \Big) + \mathbf{b}'.
\end{align}
Finally, by definition of $\Delta \matr{b}'_i(\matr{C})$, we obtain
\begin{align}
    \matr{T}_{\matr{W}_i(\matr{C}), \mathbf{b}_i'(\matr{C})}(\mathbf{x}) 
    &= (\matr{C}, \mathbf{x})_{(i)} + \matr{A}(\matr{C}, \mathbf{x})_{(i)} + \matr{W}'\sigma \Big( \matr{W} \big( \matr{A}(\matr{C}, \mathbf{x})_{(i)} + (\matr{C}, \mathbf{x})_{(i)} \big) + \mathbf{b} \Big) + \mathbf{b}' \\
    &= \matr{T}_{\matr{W}, \mathbf{b}'}(\matr{C}, \mathbf{x})_{(i)}
\end{align}
The last step is to extend, following \S \ref{any-block-result}, the result to any contextual block $\ell$
\begin{equation}
    \matr{T}^\ell(\matr{C}_\ell, \mathbf{x}_\ell)_{(i)} = (\matr{C}_\ell, \mathbf{x}_\ell)_{(i)} + \matr{A}^\ell(\matr{C}_\ell, \mathbf{x}_\ell)_{(i)} + \matr{W}'\sigma \Big( \matr{W} \big( \matr{A}^\ell(\matr{C}_\ell, \mathbf{x}_\ell)_{(i)} + (\matr{C}_\ell, \mathbf{x}_\ell)_{(i)} \big) + \mathbf{b} \Big) + \mathbf{b}'
    \label{eq:correct-skips-any-pos-any-block}
\end{equation}
where we simply index the block $\matr{T}^\ell$, layer $\matr{A}^\ell$ and input $(\matr{C}_\ell, \mathbf{x}_\ell)$, with $(\matr{C}_1, \mathbf{x}_1)$ as the original context and query. The parameter updates now become
\begin{align}
    \Delta \matr{W}_i(\matr{C}_\ell) &= \frac{\big( \matr{W} (\Delta \matr{A}^\ell_{(i)} + \Delta \mathbf{z}^\ell_{(i)}) \big) \big( \matr{A}^\ell(\mathbf{x}_\ell) + \mathbf{x}_\ell \big)^T}{||\matr{A}^\ell(\mathbf{x}_\ell) + \mathbf{x}_\ell||^2} \label{eq:skips-weight-update} \quad \text{and} \\
    \Delta \matr{b}'_i(\matr{C}_\ell) &= \Delta \matr{A}^\ell_{(i)} + \Delta \mathbf{z}^\ell_{(i)},
    \label{eq:skips-bias-update}
\end{align}
with $\Delta \mathbf{z}^\ell_{(i)} = (\matr{C}_\ell, \mathbf{x}_\ell)_{(i)} - \mathbf{x}_\ell$. By exactly the same computation as above, this leads to
\begin{equation}
    \matr{T}^\ell_{\matr{W}_i(\matr{C}_\ell), \mathbf{b}_i'(\matr{C}_\ell)}(\mathbf{x}_\ell) = \matr{T}^\ell_{\matr{W}, \mathbf{b}'}(\matr{C}_\ell, \mathbf{x}_\ell)_{(i)}
\end{equation}
which proves Theorem \ref{thm1} without layer normalisation.

\subsubsection{Generalisation to any block with skips and layer normalisations (Theorem \ref{thm1})}
\label{thm1-proof}

Building on the previous results in \S \ref{any-seq-pos-result}-\ref{more-accurate-skips-results}, here we prove Theorem \ref{thm1} in its most general form, namely for any token and block including both skip connections and Pre-LN. Omitting the layer index $\ell$ for simplicity, the Pre-LN contextual block is given by
\begin{align}
    \matr{T}(\matr{C}, \mathbf{x})_{(i)} = (\matr{C}, \mathbf{x})_{(i)} &+ \matr{A}\big( \LN(\matr{C}, \mathbf{x}) \big)_{(i)} \notag \\ 
    &+ \matr{W}'\sigma \Big( \matr{W} \LN' \big[\matr{A} \big( \LN (\matr{C}, \mathbf{x}) \big)_{(i)} + (\matr{C}, \mathbf{x})_{(i)} \big] + \mathbf{b} \Big) + \mathbf{b}'
    \label{eq:pre-ln-block}
\end{align}
where recall that layer normalisation (LN) \cite{ba2016layer} of some input vector $\mathbf{x}$ is given by $\LN(\mathbf{x}) = \boldsymbol{\gamma} \odot \frac{\mathbf{x} - \mathbb{E}[\mathbf{x}]}{\sqrt{\text{Var}[\mathbf{x}] + \epsilon}} + \boldsymbol{\beta}$, with some optional learnable factors $\boldsymbol{\gamma}$ and $\boldsymbol{\beta}$. Similar to the MLP weight matrices and biases, $\LN(\cdot)$ and $\LN'(\cdot)$ indicate the pre-attention and pre-MLP layer normalisations, respectively. As in \S \ref{more-accurate-skips-results}, we want to prove that there exist some MLP parameter updates such that
\begin{equation}
    \matr{T}^\ell_{\matr{W}, \mathbf{b}'}(\matr{C}_\ell, \mathbf{x}_\ell)_{(i)} = \matr{T}^\ell_{\matr{W} + \Delta\matr{W}_i(\matr{C}_\ell), \mathbf{b}' + \Delta \mathbf{b}_i'(\matr{C}_\ell)}(\mathbf{x}_\ell)
\end{equation}
for any token position $i$ and block $\ell$. To derive the updates, we first show some more general expressions that recover all the previous cases. Specifically, the general weight update is given by
\begin{equation}
    ( \matr{W} + \Delta \matr{W}_i ) \mathbf{f} = \matr{W} \mathbf{g}_i \implies \Delta \matr{W}_i = \frac{\matr{W}(\mathbf{g}_i - \mathbf{f})\mathbf{f}^T}{||\mathbf{f}||^2}
    \label{eq:general-weight-update-formula}
\end{equation}
where $\mathbf{g}_i$ and $\mathbf{f}$ are the \textit{full input to MLP with and without context}, respectively. For example, in the case of no skip connections, Eq. \ref{eq:general-weight-update-formula} reduces to the result of \cite{dherin2025learning} (Eq. \ref{eq:dherin-result} for $i = N$) where $\mathbf{g}_i = \matr{A}(\matr{C}, \mathbf{x})_{(i)}$ and $\mathbf{f} = \matr{A}(\mathbf{x})$, hence $\mathbf{g}_i - \mathbf{f} = \Delta \matr{A}_{(i)}$. In this case, the difference in the MLP input with and without context coincides with that of the contextual layer's output (with and without context). Note also that Eq. \ref{eq:general-weight-update-formula} shows that the implicit weight update has rank 1 for any $\mathbf{g}_i$ and $\mathbf{f}$.

If we add skip connections as in \S \ref{more-accurate-skips-results} (Eq. \ref{eq:correct-skips-any-pos}), Eq. \ref{eq:general-weight-update-formula} reduces to the derived update of Eq. \ref{eq:skips-weight-update}, where $\mathbf{g}_i = \matr{A}(\matr{C}, \mathbf{x})_{(i)} + (\matr{C}, \mathbf{x})_{(i)}$ and $\mathbf{f} = \matr{A}(\mathbf{x}) + \mathbf{x}$, hence $\mathbf{g}_i - \mathbf{f} = \Delta \matr{A}_{(i)} + \Delta \mathbf{z}_{(i)}$. Note that now the input to the MLP with and without context no longer coincides with that of the contextual layer's output and also includes a skip connection delta $\Delta \mathbf{z}_{(i)} = (\matr{C}, \mathbf{x})_{(i)} - \mathbf{x}$. In this case, as shown in \S \ref{more-accurate-skips-results}, we also need an implicit update for the biases of the last MLP layer
\begin{equation}
    \Delta \mathbf{b}'_i = \mathbf{q}_i - \mathbf{p}
    \label{eq:general-bias-update-formula}
\end{equation}
where $\mathbf{q}_i$ and $\mathbf{p}$ are the \textit{full output of the contextual layer} (including the skip) \textit{with and without context}, respectively. In this case, they reduce to the input to the MLP (with and without context) and the derived update of Eq. \ref{eq:skips-bias-update}, namely $\mathbf{q}_i - \mathbf{p} = \mathbf{g}_i - \mathbf{f} = \Delta \matr{A}_{(i)} + \Delta \mathbf{z}_{(i)}$.

\begin{figure}[H]
    \begin{center}
        \centerline{\includegraphics[width=0.9\textwidth]{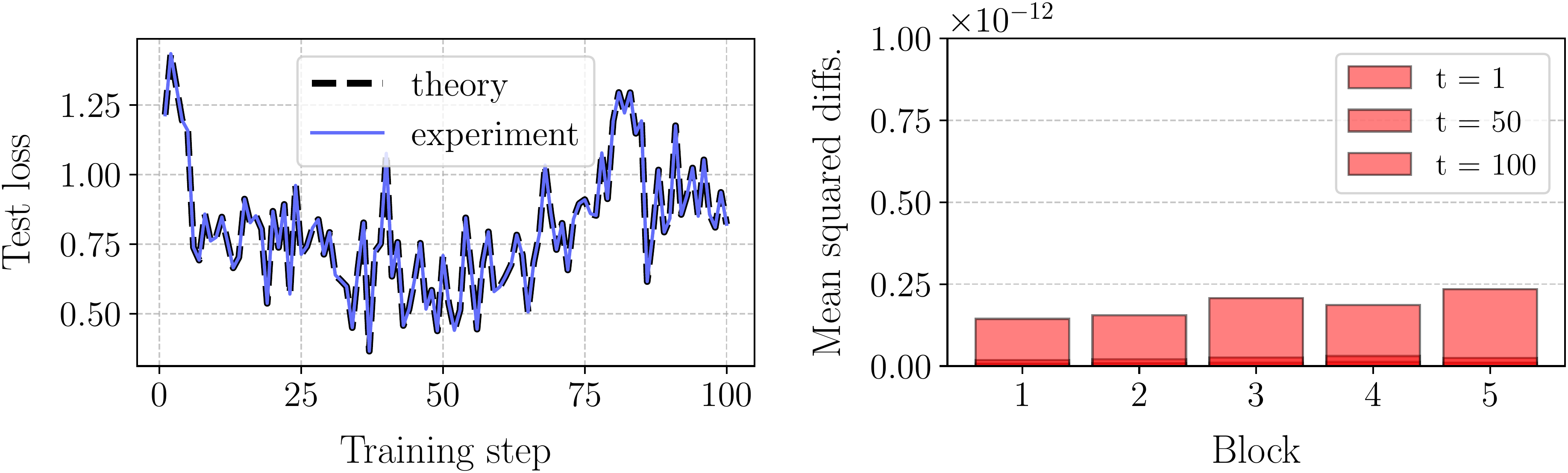}}
        \caption{\textbf{Empirical verification of Theorem \ref{thm1} for Pre-LN transformer blocks.} We plot the same metrics as in Figure \ref{fig:theory-verification} for a transformer with layer normalisation (Pre-LN), with all other hyperparameters held constant. Strangely, we found that it was more challenging to obtain good generalisation performance on in-context linear regression tasks with LN for many different hyperparameters. However, it should be noted that the Pre-LN architecture remains the standard for most large language models.}
        \label{fig:layer-norm}
    \end{center}
    \vskip -0.25in
\end{figure}

Finally, if we consider a Pre-LN contextual block as in Eq. \ref{eq:pre-ln-block}, the general weight update of Eq. \ref{eq:general-weight-update-formula} leads to 
\begin{equation}
    \Delta \matr{W}_i(\matr{C}) = \frac{\Big( \matr{W} \big(  \overbrace{\LN'\big[ \matr{A}\big( \LN(\matr{C}, \mathbf{x}) \big)_{(i)} + (\matr{C}, \mathbf{x})_{(i)} \big]}^{\mathbf{g}_i} - \overbrace{\LN'\big[ \matr{A}\big( \LN(\mathbf{x}) \big) + \mathbf{x} \big]}^{\mathbf{f}} \big) \Big) \big( \overbrace{\LN'\big[ \matr{A} \big( \LN(\mathbf{x}) \big) + \mathbf{x} \big]}^{\mathbf{f}} \big)^T}{\big|\big|\underbrace{\LN'\big[\matr{A}\big( \LN(\mathbf{x}) \big) + \mathbf{x} \big]}_{\mathbf{f}}\big|\big|^2}
    \label{eq:weight-update-ln}
\end{equation}
where note that now the difference in the full input to the MLP $\mathbf{g}_i - \mathbf{f}$ (including the input skip and LNs) does not simplify because of the nonlinear, nested LNs. The general update for the last layer's biases of Eq. \ref{eq:general-bias-update-formula} gives 
\begin{align}
    \Delta \mathbf{b}'_i(\matr{C}) &= \big[\underbrace{\matr{A}\big( \LN(\matr{C}, \mathbf{x}) \big)_{(i)} + (\matr{C}, \mathbf{x})_{(i)}}_{\mathbf{q}_i}\big] - \big[\underbrace{\matr{A}\big( \LN(\mathbf{x}) \big) + \mathbf{x}}_{\mathbf{p}}\big] \notag \\
    &= \Delta \matr{A}_{(i)} + \Delta \mathbf{z}_{(i)}.
    \label{eq:bias-update-ln}
\end{align}
Note (i) that now $\mathbf{q}_i \neq \mathbf{g}_i$ and $\mathbf{p} \neq \mathbf{f}$ because of the second (pre-MLP) LN, and (ii) that the update is the same as that without LN except that the difference in the contextual layer's output now includes the first (pre-attention) LN, i.e. $\Delta \matr{A}_{(i)} = \matr{A}\big( \LN(\matr{C}, \mathbf{x}) \big)_{(i)} - \matr{A}\big( \LN(\mathbf{x}) \big)$. Using these updates (Eqs. \ref{eq:weight-update-ln}-\ref{eq:bias-update-ln}), it can be shown by direct computation as in \S \ref{more-accurate-skips-results} that
\begin{equation}
    \matr{T}^\ell_{\matr{W}, \mathbf{b}'}(\matr{C_\ell}, \mathbf{x}_\ell)_{(i)} = \matr{T}^\ell_{\matr{W} + \Delta\matr{W}_i(\matr{C}_\ell), \mathbf{b}' + \Delta \mathbf{b}_i'(\matr{C}_\ell)}(\mathbf{x}_\ell)
\end{equation}
for any token position $i$ and block $\ell$, which concludes the proof. This particular equality for Pre-LN blocks is empirically verified in Figure \ref{fig:layer-norm}. Note that, as in \S \ref{any-seq-pos-result}, we can rewrite the result more compactly in matrix-vector form:
\begin{equation}
    \matr{T}^\ell_{\matr{W}, \mathbf{b}'}(\matr{C}_\ell, \mathbf{x}_\ell) = \matr{T}^\ell_{\matr{B} + \Delta \matr{B}(\matr{C}_\ell), \mathbf{e} + \Delta \mathbf{e}(\matr{C}_\ell)}(\mathbf{x}_\ell)
\end{equation}
where the stacked weight matrices $\matr{B}$ and their updates $\Delta \matr{B}(\matr{C}_\ell)$ are given in Eq. \ref{eq:stacked-weight-matrices} for the first block but can be similarly extended to any block, while all the biases $\mathbf{e}$ and their updates $\Delta \mathbf{e}(\matr{C}_\ell)$ are concatenated as follows
\begin{align}
    \mathbf{e}
    = \left(\begin{array}{c}
         \mathbf{b}' \\
         \mathbf{b}' \\
         \vdots \\
         \mathbf{b}'
    \end{array}\right) \in \mathbb{R}^{hN} \quad \text{and} \quad \Delta \mathbf{e}(\matr{C}_\ell) = 
    \left(\begin{array}{c}
         \Delta \mathbf{b}_1(\matr{C}_\ell) \\
         \Delta \mathbf{b}_2(\matr{C}_\ell) \\
         \vdots \\
         \Delta \mathbf{b}_N(\matr{C}_\ell)
    \end{array}\right) \in \mathbb{R}^{hN}.
    \label{eq:stacked-biases}
\end{align}
Note that $\Delta \matr{B}(\matr{C}_\ell)$ retains rank 1.

\subsection{Experimental details} 
\label{exp-details}
We trained transformers to learn linear regression tasks in context, following \cite{dherin2025learning, garg2022can, zhang2024trained}. This involved exposing the model to a sequence of input-output pairs from linear functions at training time, and testing it on previously unseen linear functions at inference time. An example sequence consists of $(\mathbf{x}_1, h_b(\mathbf{x}_1), \dots, \mathbf{x}_{N-1}, h_b(\mathbf{x}_{N-1}), \mathbf{x}_{\text{query}})$, where $\mathbf{x}_i \sim \mathcal{N}(0, \matr{I}_{d_x}) \in \mathbb{R}^{d_x}$ and $h_b(\mathbf{x}_i) = \langle \mathbf{w}, \mathbf{x}_i \rangle$ with $\mathbf{w} \sim \mathcal{N}(0, \matr{I}_{d_x}) \in \mathbb{R}^{d_x}$. Note that one function (or task) $b$ is sampled for every $N$ inputs. For input to the model, all the input-output pairs are concatenated along the $d_x$ dimension as in \cite{dherin2025learning}, such that the input or embedding matrix is $(\matr{C}, \mathbf{x}) \in \mathbb{R}^{(d_x+1) \times N}$, with $(\matr{C}, \mathbf{x})_{(N)} = [\mathbf{x}_{\text{query}}, 0]^T$.\footnote{As a small side note, \cite{dherin2025learning} write the context as having length $N$, leading to a $N+1$ sequence. We use an $N-1$ context to keep the notation compact when indexing the last token.} Transformers were trained to minimise the last-token prediction error over a batch of tasks 
\begin{equation}
    \mathcal{L}(\boldsymbol{\theta}) = \frac{1}{2B}\sum_{b=1}^B||y_b - f_{\boldsymbol{\theta}}(\matr{C}, \mathbf{x})_{(b, d, N)} ||^2
    \label{eq:mse-loss}
\end{equation}
where $y_b = h_b(\mathbf{x_{\text{query}}})$ and $\hat{y}_b = f_{\boldsymbol{\theta}}(\matr{C}, \mathbf{x})_{(b, d, N)}$ indicates the model prediction of the last token over the last input (target) dimension.

For the results of Figure \ref{fig:theory-verification}, we used batch size $B = 128$, sequence length $N = 51$ and input dimension $d_x = 2$. The transformers had $L=5$ residual blocks, each composed of a causal attention layer with 3 heads followed by a standard 2-layer MLP with GeLU as activation function. All models were trained for 100 steps using Adam \cite{kingma2014adam} with learning rate $\eta = 5e^{-2}$. The mean squared differences (MSDs) reported in Figure \ref{fig:theory-verification} were computed using 
\begin{equation}
    \MSD = \frac{1}{BNd} \sum_{b=1}^B\sum_{i=1}^N||\matr{T}^\ell_{\matr{W}, \mathbf{b}'}(\matr{C}_\ell, \mathbf{x}_\ell)_{(b, i)} - \matr{T}^\ell_{\matr{W} + \Delta \matr{W}_i(\matr{C}), \mathbf{b}' + \Delta \mathbf{b}_i'(\matr{C})}(\mathbf{x}_{\ell})_{(b)}||^2
    \label{eq:mse-metric}
\end{equation}
for every block $\ell = 1, \dots, L$. This is simply a measure of the deviation of the theoretical predictions from the empirical ones averaged over $B$ batches, $N$ sequence positions and $d$ input dimensions. Every run was repeated for different random seeds to ensure consistency. Code to reproduce all the results is available at \url{https://github.com/francesco-innocenti/generalising-icl}. All experiments were run on a CPU.

\subsection{Alignment of implicit weight updates}
\label{alignment-analysis}

\begin{figure}[H]
    \begin{center}
        \centerline{\includegraphics[width=\textwidth]{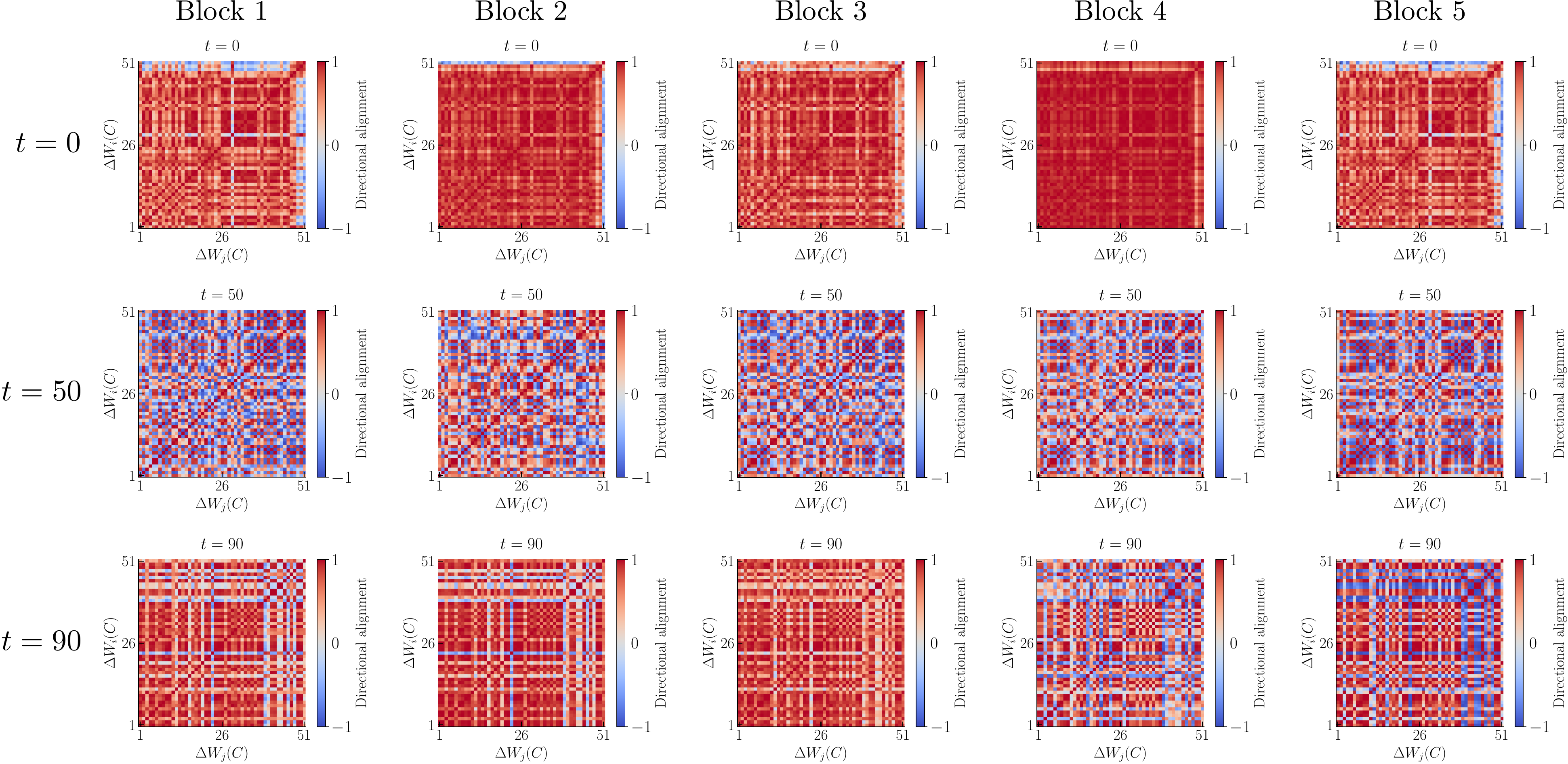}}
        \caption{\textbf{The alignment of the implicit weight updates related to different tokens has a qualitatively consistent structure across blocks.} Directional alignment (Eq. \ref{eq:directional-alignment}) between weight updates associated with different sequence positions $\DA(\Delta \matr{W}_i, \Delta \matr{W}_j)$ for $i= j = 1, \dots, N$ for each block, at different steps in training. See also Figures \ref{fig:tokens-alignment-task-8} \& \ref{fig:tokens-alignment-task-113} for other example tasks.}
        \label{fig:tokens-alignment-task-35}
    \end{center}
    \vskip -0.25in
\end{figure}

\begin{figure}[H]
    \begin{center}
        \centerline{\includegraphics[width=\textwidth]{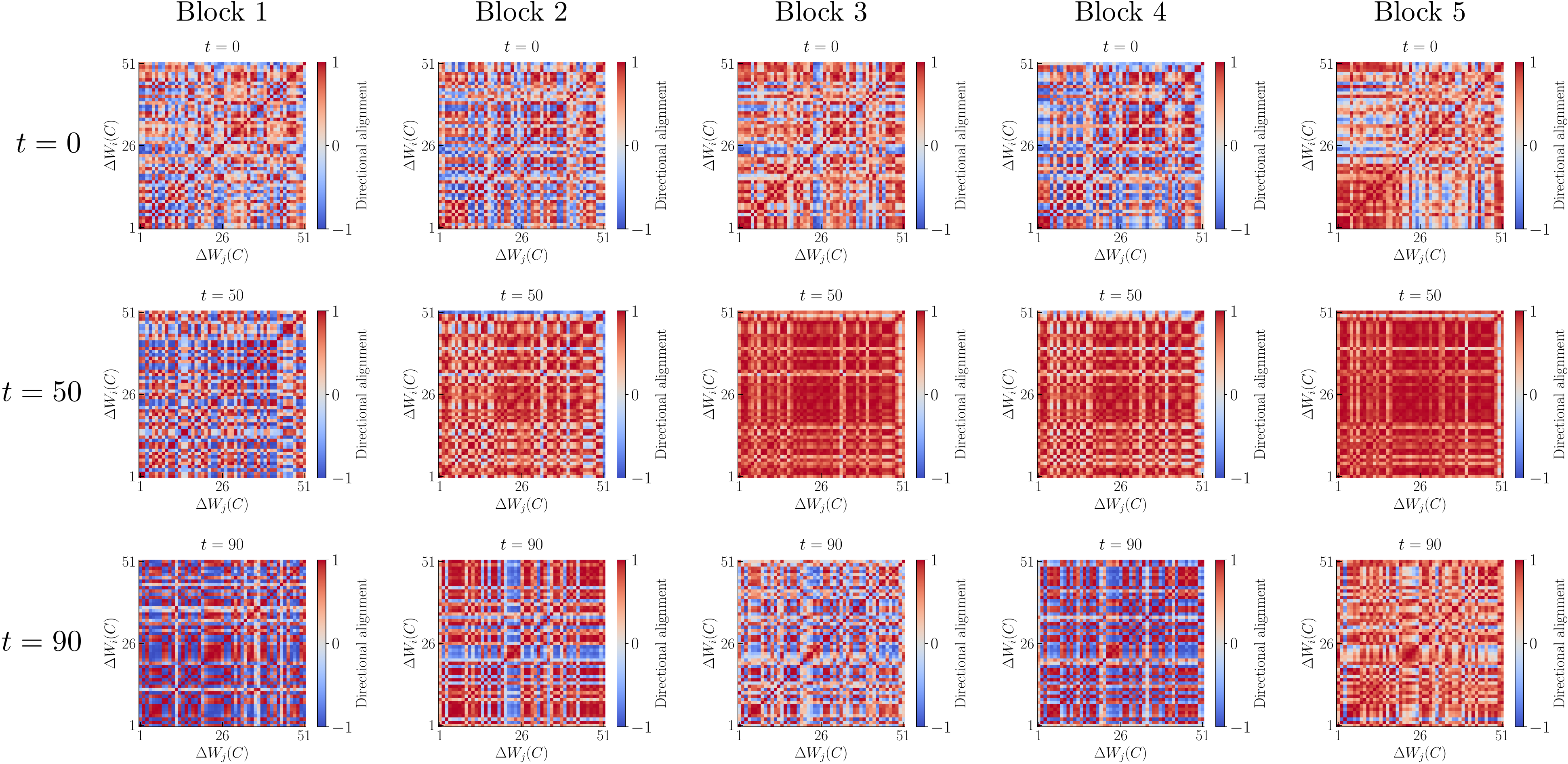}}
        \caption{\textbf{Same results as Figure \ref{fig:tokens-alignment-task-35} for a different example task or input sequence.}}
        \label{fig:tokens-alignment-task-8}
    \end{center}
    \vskip -0.25in
\end{figure}

\begin{figure}[H]
    \begin{center}
        \centerline{\includegraphics[width=\textwidth]{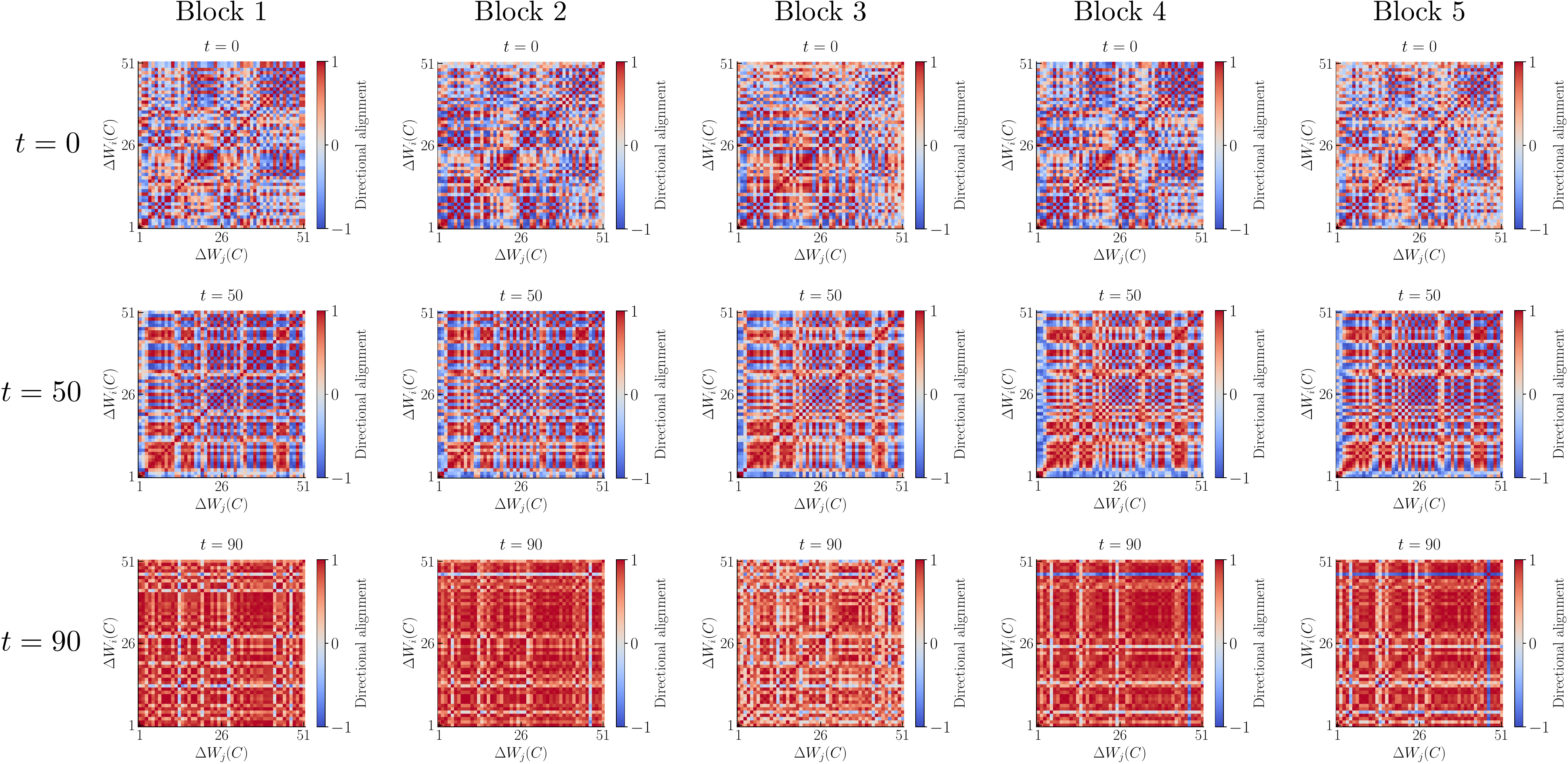}}
        \caption{\textbf{Same results as Figures \ref{fig:tokens-alignment-task-35} and \ref{fig:tokens-alignment-task-8} for yet another example task.}}
        \label{fig:tokens-alignment-task-113}
    \end{center}
    \vskip -0.25in
\end{figure}

Given our result that different token positions $i$ (as well as blocks $\ell$) are associated with different implicit weight updates (Eq. \ref{eq:main}), we investigated their relationship. The experimental setup was the same as in Figure \ref{fig:theory-verification}. As a metric of the ``directional alignment'' (DA) between any two weight updates $\Delta \matr{W}_i(\matr{C})$ and $\Delta \matr{W}_j(\matr{C})$, we computed their normalised Frobenius inner product
\begin{equation}
    \DA(\Delta \matr{W}_i, \Delta \matr{W}_j) = \frac{\langle \Delta \matr{W}_i, \Delta \matr{W}_j \rangle_F}{||\Delta \matr{W}_i||_F ||\Delta \matr{W}_j||_F},
    \label{eq:directional-alignment}
\end{equation}
where $\langle \matr{A},\matr{B} \rangle_F = \operatorname{Tr}(\matr{A}^T \matr{B})$. We first investigated the alignment between the updates related to different tokens across blocks. We find that, for a given task sequence $b$, the structure of the tokens' alignment appears qualitatively consistent across blocks (Figures \ref{fig:tokens-alignment-task-35}, \ref{fig:tokens-alignment-task-8} \& \ref{fig:tokens-alignment-task-113}).

\begin{figure}[H]
    \begin{center}
        \centerline{\includegraphics[width=0.75\textwidth]{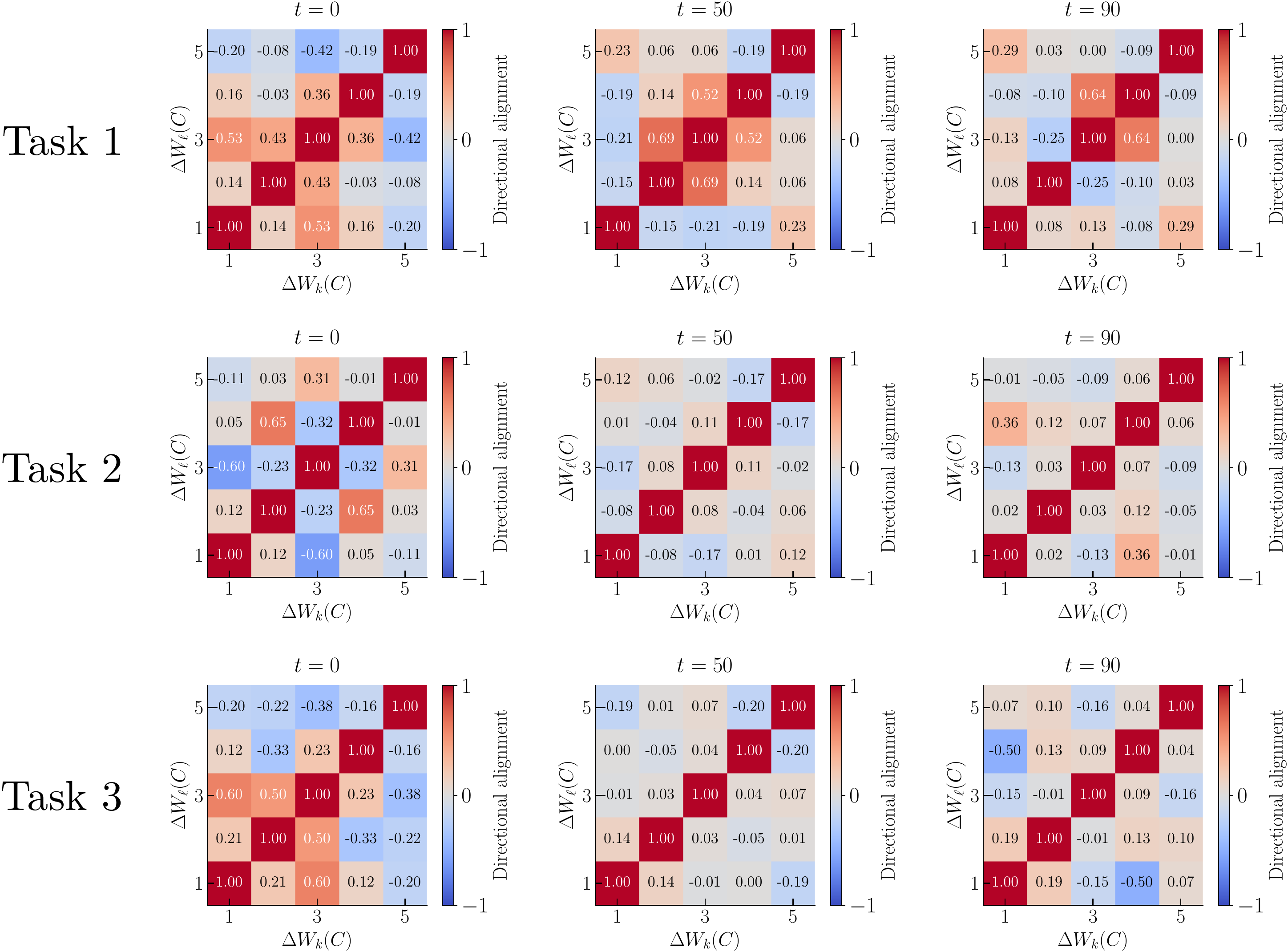}}
        \caption{\textbf{The alignment of the implicit weight update related to the last token does not share a consistent structure between blocks.} Directional alignment (Eq. \ref{eq:directional-alignment}) between weight updates associated with the last sequence element of different blocks $\DA(\Delta \matr{W}^\ell_N, \Delta \matr{W}^l_N)$ for $\ell = k = 1, \dots, L$ for different tasks $b$, at different training steps $t$. Results were consistent across different random seeds.}
        \label{fig:blocks-alignment}
    \end{center}
    \vskip -0.25in
\end{figure}

However, the alignment of the weight update related to only the last token of different blocks showed no particular structure at any point during training (Figure \ref{fig:blocks-alignment}).

\subsection{Supplementary figures} 
\label{supp-figures}
\begin{figure}[H]
    \begin{center}
        \centerline{\includegraphics[width=\textwidth]{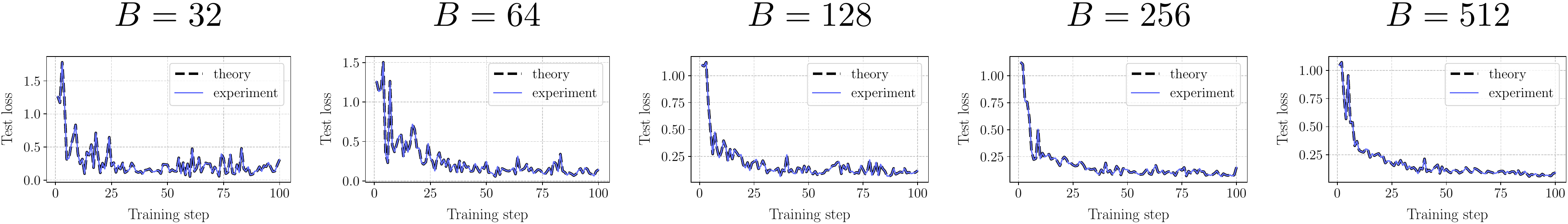}}
        \caption{\textbf{Increasing the number of tasks $B$ makes learning easier.} Empirical vs theoretical test losses on the same task as in Figure \ref{fig:theory-verification}, varying the number of tasks $B$ (i.e. number of sequences of linear functions), while holding all other hyperparameters constant.}
        \label{fig:varying-tasks}
    \end{center}
    \vskip -0.25in
\end{figure}
\begin{figure}[H]
    \begin{center}
        \centerline{\includegraphics[width=0.75\textwidth]{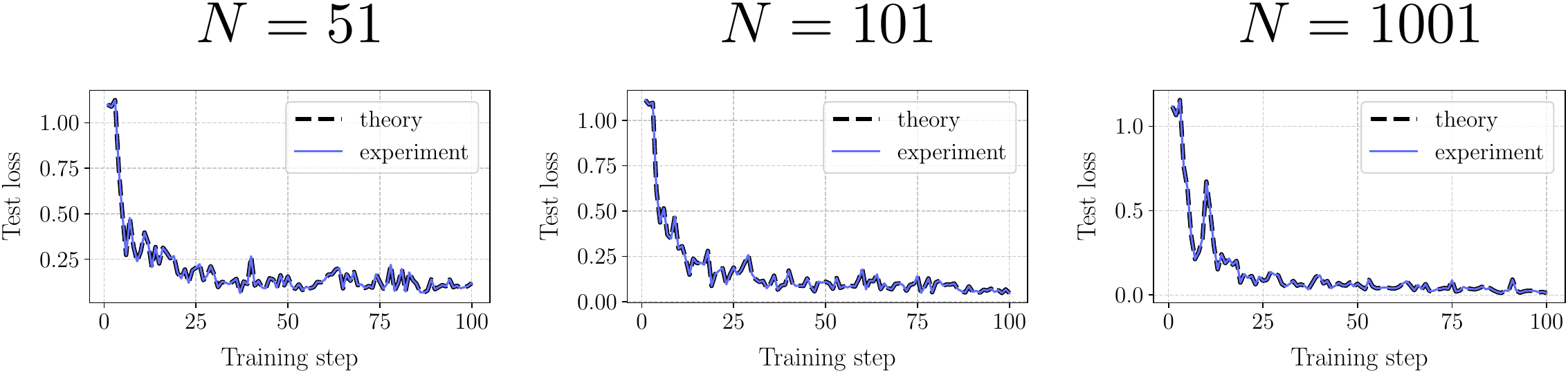}}
        \caption{\textbf{Increasing the input sequence length $N$ facilitates learning.} Empirical vs theoretical test losses on the same task as in Figure \ref{fig:theory-verification}, varying the data sequence length $N$, while holding all other hyperparameters constant.}
        \label{fig:varying-seq-len}
    \end{center}
    \vskip -0.25in
\end{figure}
\begin{figure}[H]
    \begin{center}
        \centerline{\includegraphics[width=0.75\textwidth]{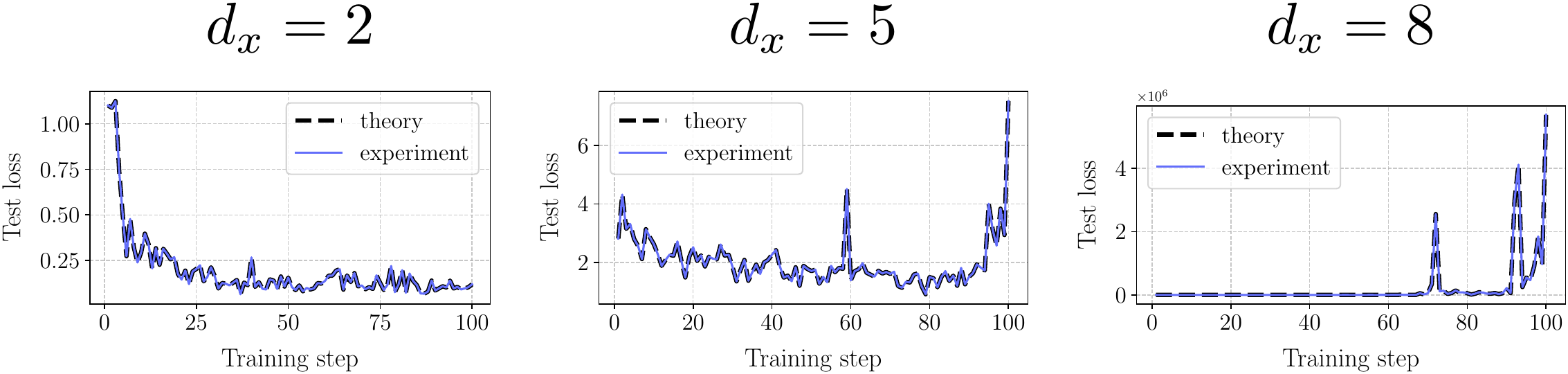}}
        \caption{\textbf{Increasing the input dimensionality $d_x$ makes learning more challenging.} Empirical vs theoretical test losses on the same task as in Figure \ref{fig:theory-verification}, varying the input dimension $d_x$ (i.e. number of regression coefficients), while holding all other hyperparameters constant.}
        \label{fig:varying-input-dim}
    \end{center}
    \vskip -0.25in
\end{figure}


\newpage
\section*{NeurIPS Paper Checklist}

\begin{enumerate}

\item {\bf Claims}
    \item[] Question: Do the main claims made in the abstract and introduction accurately reflect the paper's contributions and scope?
    \item[] Answer: \answerYes{} 
    \item[] Justification: Our claims are clearly stated in the abstract and introduction and are verified by experiments.

\item {\bf Limitations}
    \item[] Question: Does the paper discuss the limitations of the work performed by the authors?
    \item[] Answer: \answerYes{} 
    \item[] Justification: The main limitation of our work is stated in the conclusion.

\item {\bf Theory assumptions and proofs}
    \item[] Question: For each theoretical result, does the paper provide the full set of assumptions and a complete (and correct) proof?
    \item[] Answer: \answerYes{} 
    \item[] Justification: Complete proofs are provided in Appendix \ref{appendix}.

    \item {\bf Experimental result reproducibility}
    \item[] Question: Does the paper fully disclose all the information needed to reproduce the main experimental results of the paper to the extent that it affects the main claims and/or conclusions of the paper (regardless of whether the code and data are provided or not)?
    \item[] Answer: \answerYes{} 
    \item[] Justification: We provide details needed to reproduce all the experimental results in \S \ref{exp-details}.

\item {\bf Open access to data and code}
    \item[] Question: Does the paper provide open access to the data and code, with sufficient instructions to faithfully reproduce the main experimental results, as described in supplemental material?
    \item[] Answer: \answerYes{} 
    \item[] Justification: Code used to reproduce all the experimental results will be released upon publication of this work.

\item {\bf Experimental setting/details}
    \item[] Question: Does the paper specify all the training and test details (e.g., data splits, hyperparameters, how they were chosen, type of optimizer, etc.) necessary to understand the results?
    \item[] Answer: \answerYes{} 
    \item[] Justification: We specify important details needed to reproduce and understand the experiments in \S \ref{exp-details}.

\item {\bf Experiment statistical significance}
    \item[] Question: Does the paper report error bars suitably and correctly defined or other appropriate information about the statistical significance of the experiments?
    \item[] Answer: \answerNo{} 
    \item[] Justification: Where relevant, we do not report error bars because results did not significantly vary across different random seeds or runs, as we state in the captions of relevant figures.

\item {\bf Experiments compute resources}
    \item[] Question: For each experiment, does the paper provide sufficient information on the computer resources (type of compute workers, memory, time of execution) needed to reproduce the experiments?
    \item[] Answer: \answerYes{} 
    \item[] Justification: All of our experiments were run on a single CPU, as stated in \S \ref{exp-details}.
    
\item {\bf Code of ethics}
    \item[] Question: Does the research conducted in the paper conform, in every respect, with the NeurIPS Code of Ethics \url{https://neurips.cc/public/EthicsGuidelines}?
    \item[] Answer: \answerYes{} 
    \item[] Justification:

\item {\bf Broader impacts}
    \item[] Question: Does the paper discuss both potential positive societal impacts and negative societal impacts of the work performed?
    \item[] Answer: \answerNA{} 
    \item[] Justification: We see no potential positive or negative societal impact of the work since the models tested are too simple for modern AI applications.
    
\item {\bf Safeguards}
    \item[] Question: Does the paper describe safeguards that have been put in place for responsible release of data or models that have a high risk for misuse (e.g., pretrained language models, image generators, or scraped datasets)?
    \item[] Answer: \answerNA{} 
    \item[] Justification: 

\item {\bf Licenses for existing assets}
    \item[] Question: Are the creators or original owners of assets (e.g., code, data, models), used in the paper, properly credited and are the license and terms of use explicitly mentioned and properly respected?
    \item[] Answer: \answerNA{} 
    \item[] Justification:

\item {\bf New assets}
    \item[] Question: Are new assets introduced in the paper well documented and is the documentation provided alongside the assets?
    \item[] Answer: \answerNA{} 
    \item[] Justification:

\item {\bf Crowdsourcing and research with human subjects}
    \item[] Question: For crowdsourcing experiments and research with human subjects, does the paper include the full text of instructions given to participants and screenshots, if applicable, as well as details about compensation (if any)? 
    \item[] Answer: \answerNA{} 
    \item[] Justification:

\item {\bf Institutional review board (IRB) approvals or equivalent for research with human subjects}
    \item[] Question: Does the paper describe potential risks incurred by study participants, whether such risks were disclosed to the subjects, and whether Institutional Review Board (IRB) approvals (or an equivalent approval/review based on the requirements of your country or institution) were obtained?
    \item[] Answer: \answerNA{} 
    \item[] Justification:

\item {\bf Declaration of LLM usage}
    \item[] Question: Does the paper describe the usage of LLMs if it is an important, original, or non-standard component of the core methods in this research? Note that if the LLM is used only for writing, editing, or formatting purposes and does not impact the core methodology, scientific rigorousness, or originality of the research, declaration is not required.
    \item[] Answer: \answerNA{} 
    \item[] Justification:

\end{enumerate}

\end{document}